\documentclass[10pt]{article}
\usepackage{GECCO_proceedings2e,float}

\usepackage{amsmath} 
\usepackage{graphicx}
\usepackage{dcolumn}	

\newcolumntype{d}[1]{D{.}{.}{#1}}
\newcolumntype{.}{D{.}{.}{-1}}

\floatstyle{plain}
\newfloat{algorithm}{tbp}{loa}
\floatname{algorithm}{Algorithm}

\title{Potholes on the Royal Road
\thanks{\hspace{1ex}To appear in Lee Spector et al., editors, GECCO
2001: Proceedings of the Genetic and Evolutionary Computation
Conference. Morgan Kaufmann, San Francisco, 2001.}
}

\author{
	Theodore C. Belding \\ 
	Center for the Study of Complex Systems \\
	University of Michigan \\ 
	Ann Arbor, MI 48109-1120 USA \\
	\texttt{Ted.Belding@umich.edu} \\ 
	\texttt{http://www-personal.umich.edu/{\textasciitilde}streak/}
}

\begin{document}

\maketitle

\pagestyle{plain}
\thispagestyle{plain}

\begin{abstract} 
It is still unclear how an evolutionary algorithm (EA) searches a
fitness landscape, and on what fitness landscapes a particular EA will
do well.  The validity of the building-block hypothesis, a major tenet
of traditional genetic algorithm theory, remains controversial despite
its continued use to justify claims about EAs.  This paper outlines a
research program to begin to answer some of these open questions, by
extending the work done in the royal road project. The short-term goal
is to find a simple class of functions which the simple genetic
algorithm optimizes better than other optimization methods, such as
hillclimbers.  A dialectical heuristic for searching for such a class
is introduced. As an example of using the heuristic, the simple
genetic algorithm is compared with a set of hillclimbers on a simple
subset of the hyperplane-defined functions, the pothole functions.
\end{abstract}


\section{BACKGROUND}

\emph{Evolutionary algorithms} (EAs) are computational search methods
based on biological evolution.  Some common EAs are \emph{genetic
algorithms} (GAs)~\cite{holland:1992,goldberg:1989,mitchell:1996},
\emph{evolutionary programming} (EP)~\cite{fogel:1995},
\emph{evolution strategies} (ESs)~\cite{schwefel:1995}, \emph{genetic
programming} (GP)~\cite{koza:1992}, and \emph{classifier
systems}~\cite{holland:etal:1986}. The study of EAs is called
\emph{evolutionary computation} (EC).

EAs are increasingly important in such areas as function optimization,
machine learning, and modeling.  However, as Mitchell et
al. emphasized in the royal road (RR)
papers~\cite{mitchell:etal:1992,forrest:mitchell:1993b,mitchell:etal:1994},
it is still unclear how an EA searches a fitness landscape, or even
what an EA \emph{does}. It is also unknown what types of problem are
easy or hard for a particular EA, how various landscape features
affect problem difficulty for an EA, or under what circumstances an EA
will outperform another search method.  Even less work has been done
to classify the features of real-world problems that may be relevant
to EA performance.  Moreover, the selection of EA parameters such as
mutation rate or population size is still largely a black art, despite
some promising research in this area.  This lack of theory makes the
selection and configuration of an EA for a given problem difficult.

A major open question in EC is the function and importance of the
\emph{crossover} operator, which recombines two individuals.
Holland~\cite{holland:1992} has argued that crossover is central to an
EA's efficacy. The theoretical basis for this is the
\emph{building-block hypothesis} (BBH)~\cite{goldberg:1989}, which
states that an EA uses crossover to repeatedly combine compact
subsolutions with high observed fitness from different individuals,
forming more complete subsolutions with even higher observed fitness.
Such subsolutions are called \emph{building blocks}.  When an EA uses
crossover on symbol strings from the set $A^{\ell}$, where $A$ is the
set of possible symbols and $\ell$ is the length of a string, the
building blocks are short \emph{schemata} with high observed
fitness. Schemata are members of the set $(A \cup
\{\texttt{*}\})^{\ell}$, where $\texttt{*}$ is a wildcard symbol. They
are hyperplanes in the search space.  Holland's schema
theorem~\cite{holland:1992} implies that short schemata with
consistently above-average observed fitness tend to increase
exponentially in frequency over several generations.  (If operators
other than reproduction are neglected, this is true for all partitions
of the search space, not just for a partition into schemata. However,
applying crossover to symbol strings induces a natural partition of
the space into schemata~\cite{vose:1993}. Furthermore, short schemata
are preserved by crossover. This makes schemata particularly relevant
when studying EAs that use crossover on strings.)  Applying crossover
to individuals with high fitness is a plausible heuristic for
generating offspring that will also be highly fit.  The chance that
this heuristic succeeds can be quantified using Price's covariance and
selection theorem~\cite{altenberg:1995}.  Implicit in the BBH is also
the hypothesis that there are many real-world problems amenable to
solution by this process.

A common misconception is that a schema has a unique, well-defined
fitness, which is the average fitness of all of its possible
instances, and that observed fitnesses are estimates of these
``actual'' values.  In general, no such unique schema fitness exists,
and the schema theorem makes no such assumption~\cite{holland:1991}:
The observed fitness of a schema is the average fitness of its
instances in the current population.  This value depends on the
distribution of schemata in the current population, which is biased by
the EA over time.  A uniform distribution is only seen immediately
after generating an initial random population, if ever. Hence, there
is no justification for arbitrarily defining a schema's fitness to be
the average over a uniform distribution. A schema may be a highly-fit
building block in one population but not in another, even under the
same fitness function. Grefenstette~\cite{grefenstette:1993} made
essentially the same point when he criticized the ``static
building-block hypothesis''. 

The BBH is often used to explain how EAs work and to justify the
importance of crossover.  However, there is no theory that specifies
in detail the conditions necessary for the BBH to be valid and thus for
crossover to be beneficial.  While there is empirical evidence in
favor of the BBH~\cite{holland:2000}, its validity in general for the
SGA and for other EAs using crossover remains
controversial~\cite{fogel:1995}, and the schema theorem's relevance to
EA theory has been questioned as well~\cite{vose:1993,vose:1999}.  In
particular, uniform crossover, which is more likely to break up short
building blocks than traditional crossover, is very effective on some
problems~\cite{syswerda:1989}, and it may be that on some problems
crossover acts as a macromutation operator, rather than as an operator
that recombines building blocks~\cite{jones:1995c}.  More generally,
it is not clear how to formulate a BBH that is valid for an arbitrary
EA operating on an arbitrary representation of
solutions~\cite{oreilly:oppacher:1995}.

\section{RESEARCH PROGRAM}
\label{sec:research-program}

This paper presents a research program to extend the RR
papers~\cite{mitchell:etal:1992,forrest:mitchell:1993b,mitchell:etal:1994}
in testing the validity of the BBH, focusing specifically on the
\emph{simple genetic algorithm} (SGA)~\cite{goldberg:1989}.  The SGA
is a GA that uses fitness-proportionate selection, single-point
crossover, and point mutation to evolve a single panmictic population
of bit strings, with each generation completely replacing the previous
one.  I focus on the SGA because it is a relatively simple EA with a
large theoretical literature and because many EAs descend directly or
indirectly from it. A theory developed for the SGA has a relatively
good chance of being applicable to other EAs.

As in the RR papers, I use function optimization to compare
the SGA with other search algorithms.  I do this because it is an
increasingly important application for EAs, with relatively clear
performance criteria, and because a simple fitness function is
easy to design and implement.  Also, function optimization
can be viewed as search, so theories developed for it
may be relevant to other applications of search, for instance
artificial intelligence~\cite{newell:1990} and evolutionary
biology~\cite{wright:1932}.

As De Jong~\cite{dejong:1993} pointed out, the SGA is not a function
optimizer, per se. But if the BBH is valid, the SGA should work
particularly well as an optimizer on functions rich in building blocks
that can be recombined to reach the optimum. Hence, to determine the
validity of the BBH, it would be useful to know the class of functions
on which the SGA outperforms other optimizers. Examining what makes
this class of functions particularly easy for the SGA will also help
us to predict which functions the SGA will perform well on.  One step
towards this goal is to find a simple class of functions on which the
SGA outperforms hillclimbers.  This was the goal of the RR papers and
is also the immediate goal here.  To meet this goal, the SGA should
consistently perform extremely well on the functions and outperform
hillclimbers by a wide margin, for a reasonable set of performance
criteria.  (Note that this is a different question from that addressed
by Wolpert and Macready's~\cite{wolpert:macready:1997} no free lunch
(NFL) theorem.)

This paper takes a different approach from that of the RR papers,
although the ultimate goal remains the same.  In those papers, Forrest
and Mitchell~\cite{forrest:mitchell:1993b} determined that
\emph{hitchhiking} was a major factor limiting SGA performance on the
RR functions, causing it to perform worse than the random mutation
hillclimber (RMHC). Hitchhiking occurs when detrimental or neutral
alleles increase in frequency due to the presence of nearby beneficial
alleles on the same chromosome~\cite{futuyma:1998}. This can cause
beneficial alleles at the same loci as the hitchhiking alleles to die
out in the population, preventing the SGA from finding any highly-fit
individuals that contain those alleles.  (The fact that schemata do
not have unique, well defined fitnesses is a necessary precondition
for hitchhiking.)  After identifying this problem, Mitchell et
al.~\cite{mitchell:etal:1994} investigated how to make the SGA perform
more like an \emph{idealized genetic algorithm} that was unaffected by
hitchhiking.  They developed the RR function $R4$, which reduced
hitchhiking by lowering the fitness jump from one level of the
function to the next. In effect, they made the SGA outperform the RMHC
by making the functions easier for the SGA. In contrast, I attempt to
make them harder for hillclimbers. The RR functions are easy for
hillclimbers like the RMHC because they are convex: An algorithm never
needs to go downhill in order to reach the global optimum. To make
these functions hard for hillclimbers, I add \emph{potholes} to them:
valleys in the fitness landscape that block a hillclimber's path to
the optimum~\cite{mitchell:etal:1992}.  This produces the
\emph{pothole functions}, described in
Section~\ref{sec:pothole-fcns}. (Holland proposed a class of RR
functions, described by Jones~\cite{jones:1994}, that also contained
potholes.)

Pothole functions are a very simple subset of Holland's
hyperplane-defined functions (HDFs)~\cite{holland:2000}; potholes are
examples of HDF \emph{refinements}. A long-term goal is to design a
class of pothole functions with parameters that can be varied to
select the landscape features present in a function, as well as its
overall difficulty.  An arbitrary number of functions could then be
generated with the desired characteristics by using those parameters
to define probability distributions, which in turn could be used to
choose the schemata that contribute to an individual's fitness, along
with their fitness contributions.  (In this paper, these will be
called a function's \emph{significant schemata}.) Such a class would
allow a researcher to use statistical methods to calculate the
certainty of statements about an algorithm's performance across the
entire class, while being easier to understand than more general
classes of HDFs.  However, it is not yet clear what parameters or
distributions should be used, if the goal is to describe a class of
functions that is easy for the SGA yet hard for hillclimbers.  The
immediate goal is to use hand-designed pothole functions as testbeds
to determine what regions of distribution space should be used for
randomly-generated functions.

I base my work on the RR functions because they were explicitly
designed to investigate the validity of the BBH by studying the SGA's
performance on functions rich in building blocks. The significant
schemata in a RR function are not building blocks in every population,
since their fitness depends on the current population. However, the
functions are defined so that they are building blocks in all contexts
except in the occurrence of hitchhiking, since they make only positive
fitness contributions. The functions are ``rich in building blocks''
in this sense; the pothole function $p_{1}$ described in
Section~\ref{sec:pothole-fcns} has the same characteristic. In this
paper, a schema that makes a positive fitness contribution (ignoring
the fitness contribution of other schemata) will be called a
\emph{beneficial schema}. (It is possible to define a building block
as any beneficial schema, in contrast to the definition given earlier.
This definition is related to Fisher's~\cite{fisher:1958}
\emph{average excess,} but it makes the relationship between the BBH
and the schema theorem less clear [J. H. Holland, personal
communication].)  Like the RR functions, the pothole functions are not
meant to be realistic.  Since the fitness contribution of every schema
is specified in advance, schemata can be used as tracers: They can be
related to individual landscape features, and their frequency in the
EA population can be tracked over time~\cite{mitchell:etal:1992}.
Hypotheses about the effects of various landscape features on EA
behavior can then be formulated and tested. This knowledge can then be
applied to the study of real-world functions.

\begin{algorithm}[t]
\begin{enumerate}
\item Create a function that is easy for the SGA, for some performance
criteria.
\label{item:dialectical-heuristic:begin} 
\item Use domain-specific knowledge to design a simple algorithm that
is able to optimize that function better than the SGA. If no such
algorithm can be found, or if all such algorithms incorporate
unreasonable amounts of domain-specific knowledge, go to
Step~\ref{item:dialectical-heuristic:success}.
\label{item:dialectical-heuristic:design-alg}
\item Modify the function so that it is hard for the simple algorithm
yet still easy for the SGA.  If no such function can be found, go back
to Step~\ref{item:dialectical-heuristic:begin} and start over. Otherwise,
go back to Step~\ref{item:dialectical-heuristic:design-alg}.
\item Stop --- a candidate function has been found. 
\label{item:dialectical-heuristic:success}
\end{enumerate}
\caption{A dialectical heuristic for finding a simple function that is
easy for the SGA but hard for other optimizers. (Note that this
heuristic may never succeed.)}
\label{alg:dialectial-heuristic}
\end{algorithm}

Given enough domain-specific knowledge, it is plausible that a
specialized optimization method can be designed to outperform the SGA
on \emph{any} sufficiently restricted class of functions.  (The NFL
theorem does not hold if the subset of functions being considered has
measure 0 in distribution; this is true for many subsets of interest,
in particular all countable subsets [J. H. Holland, personal
communication].)  Therefore, the issue is not whether the SGA will
outperform \emph{all} other algorithms on a given class.  Rather, it
is: How much domain-specific knowledge is it reasonable to incorporate
into an algorithm before it becomes over-specialized or too expensive
to design and implement, outweighing any performance advantage over the
SGA?  A related question is: How broad must a class of functions be
before the SGA outperforms a specialized optimizer on it?  More
generally, the RR papers suggest a \emph{dialectical heuristic} to
search for a simple function that is easy for the SGA but hard for
hillclimbers (Algorithm~\ref{alg:dialectial-heuristic}).  (Note that
``dialectic'' here simply denotes ``the existence or working of
opposing forces''~\cite{simpson:weiner:1989}.) While this heuristic is
very straightforward, it has never been articulated explicitly. Since
it is a heuristic, it may never succeed; however it is a plausible way
to search for the desired class of functions.

The remainder of this paper provides a concrete example of using this
heuristic.  The pothole function $p_{1}$ is introduced and shown to be
difficult for the RMHC but easy for the SGA.  Then a variant of the
RMHC, the lax random-mutation hillclimber (LRMHC), is defined, which
knows the depth of the potholes in $p_{1}$ and is able to cross over
them to reach the optimum.  This hillclimber is shown to outperform
the SGA on $p_{1}$.  The paper concludes with a discussion of the
results.

\begin{table*}[t]
\caption{The pothole function $p_{1}$. An individual's fitness is
calculated by summing the fitness contributions of the schemata of
which the individual is an instance, and then adding this sum to a
base fitness of 100. If the result is less than 0, it is reset to
0. The global optimum is a string of 64 \texttt{1}s, with a net
fitness of 115.}
\setlength{\arrayrulewidth}{.8pt}
\begin{center}
\footnotesize
\begin{tabular}{|r l l r|} \hline
Level & \multicolumn{2}{l}{Schema} & Fitness \\ \hline
$1$ & $s_{0}$ & \texttt{11111111********************************************************} & $1.0$ \\
 & $s_{1}$ & \texttt{********11111111************************************************} & $1.0$ \\
 & $s_{2}$ & \texttt{****************11111111****************************************} & $1.0$ \\
 & $s_{3}$ & \texttt{************************11111111********************************} & $1.0$ \\
 & $s_{4}$ & \texttt{********************************11111111************************} & $1.0$ \\
 & $s_{5}$ & \texttt{****************************************11111111****************} & $1.0$ \\
 & $s_{6}$ & \texttt{************************************************11111111********} & $1.0$ \\
 & $s_{7}$ & \texttt{********************************************************11111111} & $1.0$ \\ \hline
2 & $s_{8}$ & \texttt{1111111111111111************************************************} & $1.4$ \\
 & $s_{9}$ & \texttt{****************1111111111111111********************************} & $1.4$ \\
 & $s_{10}$ & \texttt{********************************1111111111111111****************} & $1.4$ \\
 & $s_{11}$ & \texttt{************************************************1111111111111111} & $1.4$ \\ \hline
3 & $s_{12}$ & \texttt{11111111111111111111111111111111********************************} & $1.0$ \\
 & $s_{13}$ & \texttt{********************************11111111111111111111111111111111} & $1.0$ \\ \hline
4 & $s_{14}$ & \texttt{1111111111111111111111111111111111111111111111111111111111111111} & $1.0$ \\ \hline
 & $s_{15}$ & \texttt{111111111*******************************************************} & $-0.1$ \\
 & $s_{16}$ & \texttt{11111111*1******************************************************} & $-0.1$ \\
 & $s_{17}$ & \texttt{******1*11111111************************************************} & $-0.1$ \\
 & $s_{18}$ & \texttt{*******111111111************************************************} & $-0.1$ \\
 & $s_{19}$ & \texttt{****************111111111***************************************} & $-0.1$ \\
 & $s_{20}$ & \texttt{****************11111111*1**************************************} & $-0.1$ \\
 & $s_{21}$ & \texttt{**********************1*11111111********************************} & $-0.1$ \\
 & $s_{22}$ & \texttt{***********************111111111********************************} & $-0.1$ \\
 & $s_{23}$ & \texttt{********************************111111111***********************} & $-0.1$ \\
 & $s_{24}$ & \texttt{********************************11111111*1**********************} & $-0.1$ \\
 & $s_{25}$ & \texttt{**************************************1*11111111****************} & $-0.1$ \\
 & $s_{26}$ & \texttt{***************************************111111111****************} & $-0.1$ \\
 & $s_{27}$ & \texttt{************************************************111111111*******} & $-0.1$ \\
 & $s_{28}$ & \texttt{************************************************11111111*1******} & $-0.1$ \\
 & $s_{29}$ & \texttt{******************************************************1*11111111} & $-0.1$ \\
 & $s_{30}$ & \texttt{*******************************************************111111111} & $-0.1$ \\ \hline
\end{tabular}
\normalsize
\end{center}
\label{table:p1}
\end{table*}

\section{POTHOLE FUNCTIONS}
\label{sec:pothole-fcns}

Following the dialectical heuristic described in
Section~\ref{sec:research-program}, I modified the RR functions to
make them harder for simple hillclimbers by adding \emph{potholes}.
Potholes are detrimental schemata that contain beneficial schemata,
and which, in turn, are necessary to reach beneficial schemata with
higher fitness contributions~\cite{mitchell:etal:1992}.  This produces
the class of \emph{pothole functions}. All experiments in this paper
were performed on the pothole function $p_{1}$, which is defined in
Table~\ref{table:p1}.  That table lists all of the schemata that
contribute to an individual's fitness, along with their fitness
contributions; these schemata are called the function's
\emph{significant schemata}.

The fitness $p_{1}(x)$ of a string $x \in \{0, 1\}^{64}$ is given by
\begin{equation}
p_{1}(x) = \max \left\{ 0, \  100 + \sum_{s \in S \mid x \in s} \mu(s)
\right\}, \label{eqn:pothole-fitness}
\end{equation}
where $S$ is the set of significant schemata for $p_{1}$, $s$ is a
schema in $S$, and $\mu(s)$ is the fitness contribution of $s$.  The
notation $x \in s$ stands for ``the string $x$ is an instance of the
schema $s$''. Individuals have a base fitness of 100, so that in other
pothole functions they may be less fit than the base fitness without
having a negative fitness; the fitness is forced to be equal or
greater than 0 so that fitness-proportionate selection may be
used. The global optimum is a string of 64 \texttt{1}s, which has a
fitness of 115.

The function consists of 4 \emph{levels}, defined in
Table~\ref{table:p1}.  The first level consists of \emph{elementary
schemata,} each of which is a block of 8 \texttt{1}s.  Each higher
level consists of \emph{compound schemata} composed of schemata from
the previous one. The elementary and compound schemata are all
beneficial schemata. An algorithm is said to \emph{reach} a level when
it finds an individual that is an instance of at least one significant
schema from that level.

If $p_{1}$ consisted only of schemata $s_{0}$--$s_{14}$, it would be a
RR function, similar to $R2$~\cite{mitchell:etal:1992}.  The
additional schemata $s_{15}$--$s_{31}$ are potholes.  The potholes
$s_{15}$ and $s_{16}$ prevent a single-mutation hillclimber, such as
the RMHC~\cite{forrest:mitchell:1993b}, that has reached the
first-level schema $s_{0}$ from reaching the second-level schema
$s_{8}$.  This is because every sequence of single-bit mutations that
leads from $s_{0}$ to $s_{8}$ would force the hillclimber to go
downhill in fitness through one of these potholes (assuming neither of
them is present to begin with), which it cannot do.  Similarly, the
potholes $s_{17}$ and $s_{18}$ prevent it from reaching $s_{8}$ if it
has reached $s_{1}$. The remaining potholes block the path to the
other second-level schemata.

\section{EXPERIMENTS ON $p_{1}$}
\label{sec:p1}

I first compared the SGA against a variety of hillclimbers on $p_{1}$,
to verify that it was more difficult than the RR function $R2$
for hillclimbers such as RMHC.

\subsection{SIMPLE GENETIC ALGORITHM (SGA)}

The SGA used a population of $512$ individuals.  Two offspring were
produced for each pair of parents, and the entire population was
replaced in each new generation. Standard one-point crossover was used
with a probability of $0.7$ per mating pair.  Point mutation was
applied to each offspring with a probability of $0.005$ per allele
(mutations simply flipped the allele from $0$ to $1$ or vice-versa).
Fitness-proportionate, or ``roulette wheel'', selection was used, with
$\sigma$-truncation scaling~\cite{goldberg:1989}:
\begin{equation}
f' = \max\{\min[f - (\bar{f} - c \sigma), 1.5], 0\}. 
\end{equation}
Here $f'$ is an individual's scaled fitness, $f$ is its unscaled
fitness, $\bar{f}$ is the population average unscaled fitness,
$\sigma$ is the standard deviation of unscaled fitness in the
population, and the scaling constant $c = 2$.  The maximum and minimum
possible scaled fitnesses are 1.5 and 0, respectively. The scaled
fitnesses were then used to select the parents of the next generation.
If $\sigma < 0.0001$, the unscaled fitnesses were used instead.
Scaling appears to be necessary for the SGA to do well on these
functions.

These parameters were chosen rather arbitrarily, since the goal
is to find a class of functions that the SGA can optimize easily,
without being sensitive to the exact parameter settings.

\subsection{HILLCLIMBERS}

When comparing the SGA with hillclimbers, it is important to report
results from a variety of hillclimbers.  In these experiments, I
used the next-ascent hillclimber (NAHC), steepest-ascent hillclimber
(SAHC), and random-mutation hillclimber (RMHC) described by Forrest
and Mitchell~\cite{forrest:mitchell:1993b}, and Jones's crossover
hillclimber (XOHC)~\cite{jones:1995c}. 

The XOHC used differs from Jones's in that it repeats if the maximum
number of jumps is reached, until the maximum number of evaluations
has been performed. Jones's original algorithm also quit if no fitness
increase was found within $10000$ steps; the one used here does not.

\subsection{PERFORMANCE CRITERIA}

In order for one algorithm to outperform another in this study, it
should do so over a wide range of reasonable performance metrics.  I
use the number of function evaluations needed to reach the optimum as
the primary performance metric, under the assumption that function
evaluation dominates an optimizer's running time.  When one algorithm
reaches the optimum more often than the other, I use the number of
times the optimum is reached as the primary performance metric.  When
neither algorithm reaches the optimum, I use the number of evaluations
needed to reach each level, as well as the number of times each level
was reached. The maximum fitness reached in each run is also recorded,
but not shown here.

Fitness timeseries were also plotted for each algorithm, sampled every
512 function evaluations and averaged over the set of runs; only those
for the SGA and two of the hillclimbers are shown here. (Individual
runs were also plotted but are not shown here.)  These provide much
more information about the course of each run, including the rate of
improvement in fitness.  For the SGA, the population best and average
unscaled fitness are plotted; for the hillclimbers the fitness of the
best individual evaluated so far is plotted, along with the fitness of
the current individual being evaluated.

\subsection{EXPERIMENTAL RESULTS}

\begin{figure}
\begin{center}
	\includegraphics[width=3.0in]{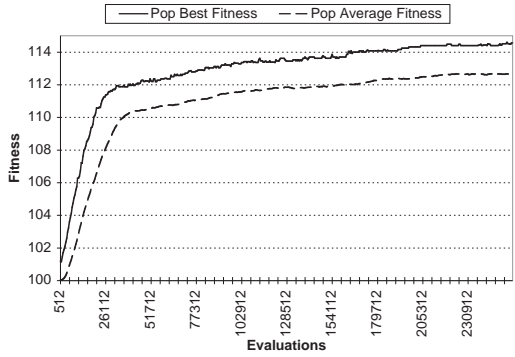}%
\end{center}
\caption{SGA population best and average fitness on $p_{1}$, sampled
every generation and averaged over 50 runs.}
\label{fig:p1:sga:timeseries}
\end{figure}

\begin{figure}
\begin{center}
	\includegraphics[width=3.0in]{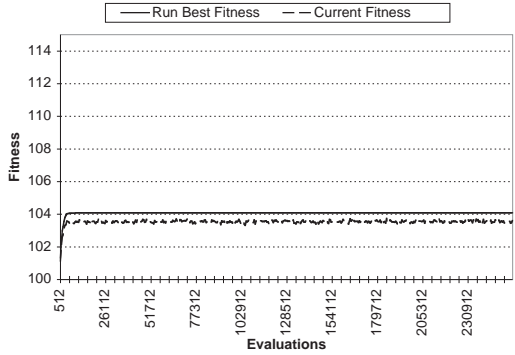}%
\end{center}
\caption{RMHC best and current fitness on $p_{1}$, sampled every 512
function evaluations and averaged over 50 runs.}
\label{fig:p1:rmhc:timeseries}
\end{figure}

The SGA, NAHC, SAHC, RMHC, and XOHC were each run 50 times on $p_{1}$,
for 256000 function evaluations per run.  The results are presented in
Table~\ref{table:p1:levels:evals}.  Timeseries for the SGA and RMHC
are presented in
Figures~\ref{fig:p1:sga:timeseries}--\ref{fig:p1:rmhc:timeseries}.
The function $p_{1}$ achieved the goal of being hard for the RMHC.
Neither it nor any of the other hillclimbers ever found the optimum.
Among these algorithms, only the SGA ever found the optimum (level 4),
and it did so in almost every run. The timeseries for the SGA and the
RMHC are consistent with the other performance metrics.

\begin{table}[t]
\caption{The SGA, RMHC, NAHC, SAHC, XOHC, and LRMHC on $p_{1}$, 50
runs, 256000 function evaluations: Mean evaluations needed to reach
each level (level 4 is the optimum). Here $n$ is the number of times a
level was reached, $\bar{x}$ is the sample mean number of evaluations
needed to reach each level, averaged over $n$, and $s$ is the sample
standard deviation of the number of evaluations.}
\setlength{\arrayrulewidth}{.8pt}
\begin{center}
\(
\begin{array}{|l l|d{1} d{1} d{1} d{1}|} \hline
 &  & \multicolumn{4}{c|}{\text{Level}} \\ 
 & & 1 & 2 & 3 & 4 \\ \hline
%
%
\text{SGA} & n & 50 & 50 & 50 & 48 \\
& \bar{x} & 36.9 & 4026.9 & 17838.2 & 65063.2 \\
& s & 32.6 & 1833.5 & 13546.4 & 50361.2 \\ \hline
\text{NAHC} & n & 50 & 48 & 0 & 0 \\
& \bar{x} & 224.6 & 72733.2 & \text{---} & \text{---} \\
& s & 219.1 & 62128.0 & \text{---} & \text{---} \\ \hline
\text{SAHC} & n & 50 & 48 & 0 & 0 \\
& \bar{x} & 213.3 & 71687.3 & \text{---} & \text{---} \\
& s & 183.1 & 57099.5 & \text{---} & \text{---} \\ \hline
%
%
\text{RMHC} & n & 50 & 2 & 0 & 0 \\
& \bar{x} & 333.7 & 3133.5 & \text{---} & \text{---} \\
& s & 289.6 & 1051.5 & \text{---} & \text{---} \\ \hline
%
%
\text{XOHC} & n & 50 & 50 & 34 & 0 \\
& \bar{x} & 392.0 & 10454.8 & 110519.0 & \text{---} \\
& s & 453.8 & 8675.6 & 63414.0 & \text{---} \\ \hline
\text{LRMHC} & n & 50 & 50 & 50 & 50 \\
 (\epsilon = 0.1) & \bar{x} & 249.9 & 1342.5 & 3547.1 & 6244.0 \\
& s & 229.0 & 915.0 & 2025.4 & 3055.2 \\ \hline
\end{array}
\)
\end{center}
\label{table:p1:levels:evals}
\end{table}

\begin{figure}
\begin{center}
	\includegraphics[width=3.0in]{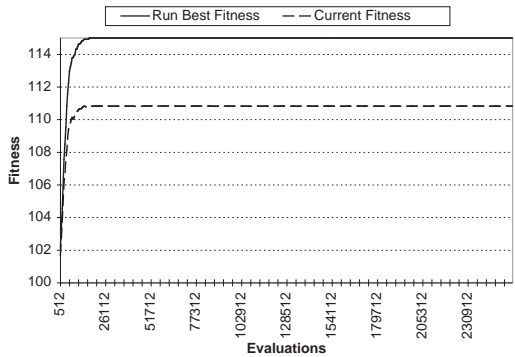}%
\end{center}
\caption{LRMHC best and current fitness on $p_{1}$, sampled every 512
function evaluations and averaged over 50 runs. $(\epsilon = 0.1)$.}
\label{wfig:p1:lrmhc:timeseries}
\end{figure}

\section{LAX RANDOM-MUTATION HILLCLIMBER (LRMHC)}

Following the dialectical heuristic presented in
Section~\ref{sec:research-program}, the next step was to see how hard
it was to design a hillclimber that outperformed the SGA on $p_{1}$.
Since the fitness penalty of each pothole was $0.1$, and the fitness
contribution of each beneficial schema was either $1.0$ or $1.4$,
there is a foolproof method for determining whether a drop in fitness
resulted from encountering a pothole, or from something else.  If the
drop is $0.1$, then it is due to just a pothole and can be ignored.
If it greater than or equal to 0.8, one or more beneficial schemata
have been lost.

\begin{algorithm}
\begin{enumerate}
\item Initialize the current individual to a random string.
\item Mutate one randomly-chosen allele.  If the new string has a
fitness equal to or greater than the current individual's fitness
minus $\epsilon$, replace the current individual with the new
individual. \label{item:lrmhc:mutate}
\item If the number of fitness evaluations performed so far is less
than the maximum, go to Step~\ref{item:lrmhc:mutate}.  Otherwise,
stop.
\end{enumerate}
\caption{The lax random-mutation hillclimber (LRMHC)
algorithm. $\epsilon$ is set to 0.1 in this paper.}
\label{alg:lrmhc-algorithm}
\end{algorithm}

The lax random-mutation hillclimber (LRMHC) listed in
Algorithm~\ref{alg:lrmhc-algorithm} resulted from incorporating this
domain-specific knowledge into the RMHC.  The LRMHC is exactly like
the RMHC, except that it accepts any new string whose fitness is no
more than $\epsilon$ below the fitness of the current string; in these
experiments, $\epsilon$ is $0.1$.  (On $p_{1}$, $\epsilon$ can be any
value in the interval $[0.1,0.8)$.) The algorithm is very similar to
the constant threshold algorithm (CTA) developed independently by
Quick et al.~\cite{quick:etal:1996}. The only difference is that the
CTA accepts a new string only if its fitness is strictly greater than
the old string's fitness minus $\epsilon$, rather than greater than or
equal as in the LRMHC. (Due to a typo, the LRMHC algorithm published
by Holland~\cite{holland:2000} also differs in this way from the
algorithm listed here.) In turn, both algorithms are similar to the
record-to-record travel algorithm and the great deluge
algorithm~\cite{dueck:1993}, and to threshold
accepting~\cite{dueck:scheuer:1990}.

In effect, the LRMHC assumes that any function it encounters has
potholes that all have a depth of no more than $\epsilon$, and that it
can ignore them since building blocks have an observed fitness
contribution higher than this value.  It might seem unreasonable to
incorporate this knowledge into the LRMHC.  But the RMHC can be viewed
as incorporating just as much knowledge; it merely assumes that the
pothole depth is always $0$. The issue here is not whether the LRMHC
is a useful general-purpose optimizer for real functions. Rather, it
is: How much domain-specific knowledge must be built into a
hillclimber so that it outperforms the SGA on $p_{1}$?

\subsection{LRMHC EXPERIMENTAL RESULTS}

Results for LRMHC on $p_{1}$ are shown in
Table~\ref{table:p1:levels:evals} and
Figure~\ref{fig:p1:rmhc:timeseries}.  It outperforms the SGA and all
of the other algorithms on $p_{1}$, always finding the optimum, and
finding it much faster than the SGA.  The timeseries for LRMHC also
shows a rapid increase in fitness over the SGA and RMHC.  These
results demonstrate that there is a very simple algorithm that
outperforms the SGA by a wide margin on $p_{1}$. Similarly, Quick et
al.~\cite{quick:etal:1996} showed that the CTA outperformed a GA
variant on a class of RR functions proposed by Holland and described
by Jones~\cite{jones:1994}, which also contained potholes.  (However,
the SGA outperforms this class of hillclimber on the RR function
$R4$~\cite{mitchell:etal:1994}.)

\section{CONCLUSIONS}

This paper has presented a research program to investigate the
validity of the BBH, as well as some preliminary results. A
dialectical heuristic for finding a simple class of functions on which
the SGA outperformed other simple search algorithms was presented.  A
class of pothole functions was designed by adding potholes to the
RR functions, in order to make them harder for simple
hillclimbers. The pothole function $p_{1}$ was shown to be hard for
hillclimbers such as the RMHC but easy for the SGA. Then a new
hillclimber, the LRMHC, was designed by incorporating domain-specific
knowledge about $p_{1}$ into the RMHC. While LRMHC is not useful as a
general-purpose optimizer, this simple hillclimber outperformed the
SGA on $p_{1}$, demonstrating that simple pothole functions such as
$p_{1}$ are still too easy for hillclimbers.  This result does not by
itself invalidate the BBH. However, it reinforces the finding of the
RR papers that simple assumptions about what functions are
especially easy for the SGA, relative to other optimizers, are often
unjustified.  Simple hillclimbers can be surprisingly effective at
optimizing simple functions.

The next step, following the dialectical heuristic from
Section~\ref{sec:research-program}, is to modify $p_{1}$ so that it
becomes hard for the LRMHC, while remaining easy for the SGA.  First,
however, the LRMHC's behavior on $p_{1}$ must be investigated, in
order to predict what kinds of functions it will find difficult.

\subsubsection*{Acknowledgments}

I thank the UM Royal Road Group, the UM Complex Systems Reading Group,
and the anonymous reviewers for their comments and suggestions.

\bibliographystyle{plain} 
\bibliography{potholes} 
\end{document}